\documentclass[sigconf]{acmart}
\usepackage{graphicx}
\usepackage{amsmath,amsfonts}
\usepackage{amsthm}
\usepackage{booktabs}
\usepackage{multirow}
\usepackage{bm}
\usepackage{subfigure}
\usepackage{xcolor}
\usepackage[linesnumbered,ruled,noend]{algorithm2e} 
\usepackage{algorithmic}
\usepackage{siunitx}
\usepackage[switch]{lineno}
\usepackage{paralist}
\usepackage{chngcntr}
\usepackage{wrapfig}
\usepackage{adjustbox}
\usepackage{siunitx}
\usepackage{paralist}
\usepackage{url}
\usepackage[skip=2pt]{caption}
\AtBeginDocument{%
  \providecommand\BibTeX{{%
    \normalfont B\kern-0.5em{\scshape i\kern-0.25em b}\kern-0.8em\TeX}}}

\copyrightyear{2021}
\acmYear{2021}
\setcopyright{acmcopyright}
\acmConference[CIKM '21] {Proceedings of the 30th ACM International Conference on Information and Knowledge Management}{November 1--5, 2021}{Virtual Event, Australia.}
\acmBooktitle{Proceedings of the 30th ACM Int'l Conf. on Information and Knowledge Management (CIKM '21), November 1--5, 2021, Virtual Event, Australia}
\acmPrice{15.00}
\acmISBN{978-1-4503-8446-9/21/11}
\acmDOI{10.1145/3459637.3481935}
\acmISBN{978-1-4503-8446-9/21/11}

\begin{document}

\title{LightMove: A Lightweight Next-POI Recommendation for Taxicab Rooftop Advertising}

\author{Jinsung Jeon}
\email{jjsjjs0902@yonsei.ac.kr}
\affiliation{%
  \institution{Yonsei University}
  \state{Seoul}
  \country{South Korea}
}

\author{Soyoung Kang}
\email{sy.kang@navercorp.com}
\affiliation{%
  \institution{NAVER Clova}
  \state{Seoul}
  \country{South Korea}
}

\author{Minju Jo, Seunghyeon Cho,\\Noseong Park}
\email{{alflsowl12,seunghyeoncho,noseong}@yonsei.ac.kr}
\affiliation{%
  \institution{Yonsei University}
  \state{Seoul}
  \country{South Korea}
}



\author{Seonghoon Kim,\\Chiyoung Song}
\email{{seonghoon.kim,chiyoung.song}@motov.co.kr}
\affiliation{%
  \institution{Motov Inc., Ltd.}
  \state{Seoul}
  \country{South Korea}
}




\renewcommand{\shortauthors}{Jeon et al.}

\begin{abstract}
Mobile digital billboards are an effective way to augment brand-awareness. Among various such mobile billboards, taxicab rooftop devices are emerging in the market as a brand new media. Motov is a leading company in South Korea in the taxicab rooftop advertising market. In this work, we present a lightweight yet accurate deep learning-based method to predict taxicabs' next locations to better prepare for targeted advertising based on demographic information of locations. Considering the fact that next POI recommendation datasets are frequently sparse, we design our presented model based on neural ordinary differential equations (NODEs), which are known to be robust to sparse/incorrect input, with several enhancements. Our model, which we call LightMove, has a larger prediction accuracy, a smaller number of parameters, and/or a smaller training/inference time, when evaluating with various datasets, in comparison with state-of-the-art models.
\end{abstract}




\maketitle

\section{Introduction}

\begin{figure}[t]
    \centering
    \subfigure[Taxicab with our rooftop advertising device]{\includegraphics[width=1\columnwidth]{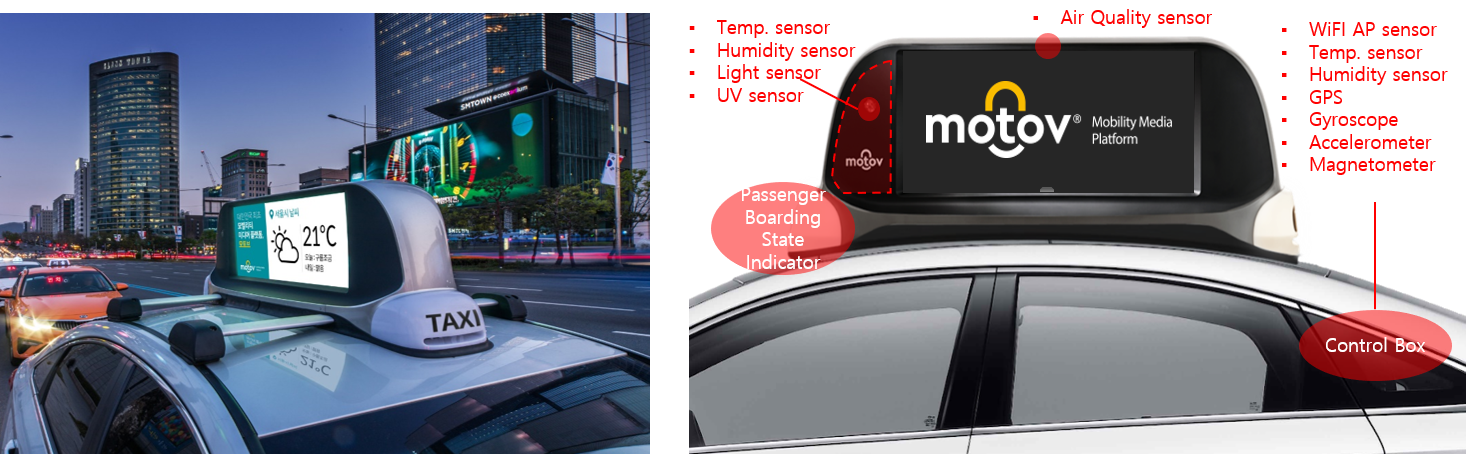}}
    \subfigure[Model Architecture]{\includegraphics[width=1\columnwidth]{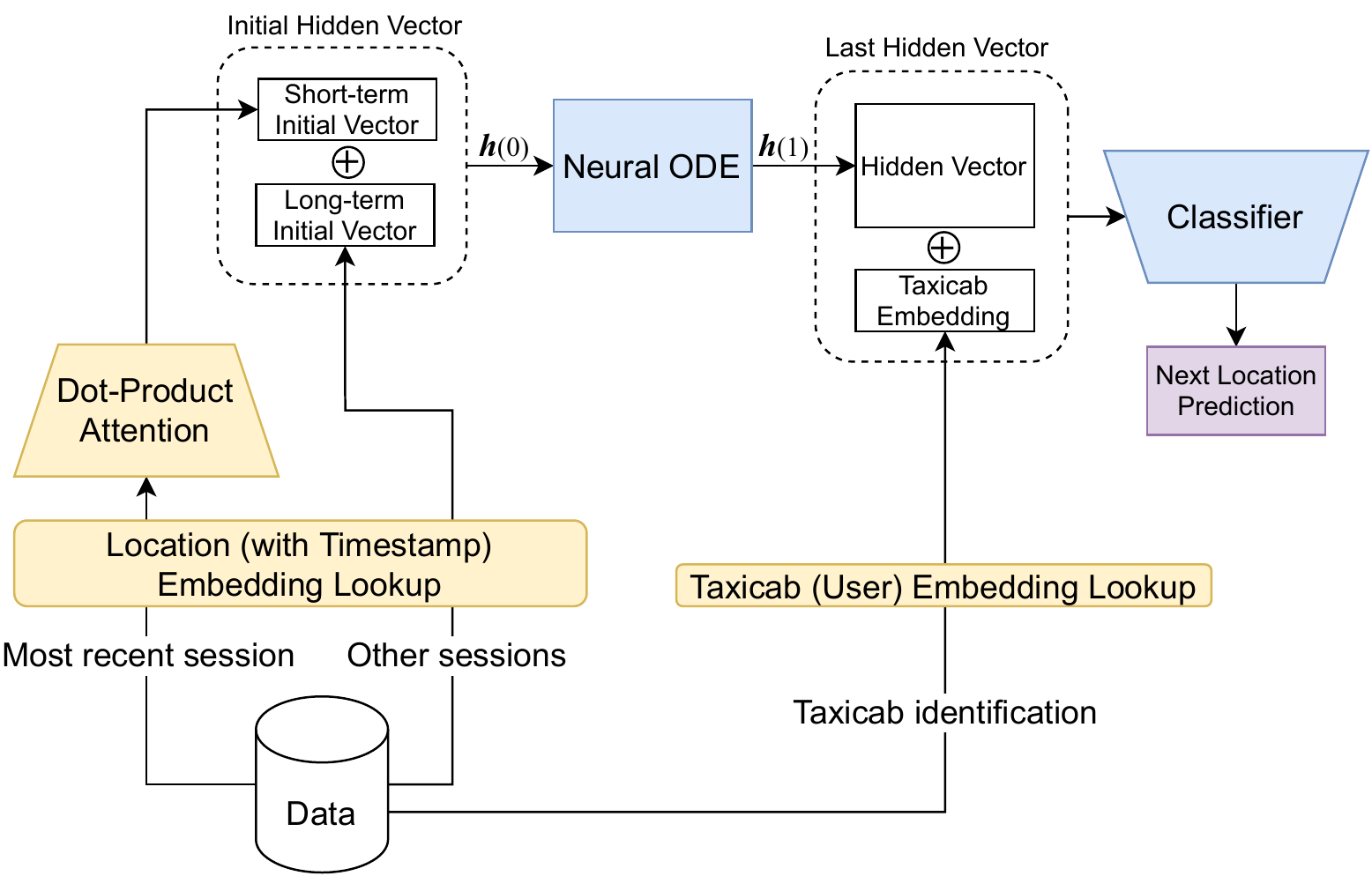}}
    \caption{Our device and prediction model. (a) Taxicab with a rooftop display (b) Our method has i) a dot-product attention for the embeddings of a set of $K$ recent locations, i.e., a recent session, ii) the embeddings for other past sessions without attention, iii) a NODE to evolve the concatenation of the short-term and the long-term embeddings, denoted $\bm{h}(0)$, to derive the hidden vector $\bm{h}(1)$ which will be concatenated with a taxicab embedding, and iv) a classifier.}
    \label{fig:archi}
\end{figure}

Mobile digital billboards promise to augment brand-awareness by targeting audiences of a specific demographic group at a particular location and time. In particular, a 2020 Nielsen report says that digital taxi rooftop billboards show more promising performance experiences than any other out of home media in the New York market~\cite{Neilson}. According to the report, watching an advertisement on a digital taxicab rooftop leads to actions for about 46\% of viewers, such as searching websites, using hashtags, downloading apps, etc. Due to such high potentiality, several start-up players like Firefly (invested by Google ventures), Uber powered by Adomni, and Motov strive to win in the market. 

Among them, Motov is a leading startup company in South Korea and is currently running about a thousand taxicab rooftops in South Korea and is supposed to increase its number up to 5,000 by the end of 2021, delivering millions of impressions per month --- an impression means a target audience is exposed once to our advertisement. In addition, Motov’s digital taxicab rooftop device is equipped with diverse sensors such as gyroscope, accelerometers, a WiFi sensor that captures surrounding pedestrians, an air quality sensor (e.g., dust density, CO2, etc.). These unique features enable advertisers to reach their target audience efficiently.

In our business, it is very important to predict the next locations of taxicabs in several minutes to prepare for targeted advertising. Given a predicted next location and its demographic statistics, we have to send image/video files to be displayed via 4G cellular network. Because taxicabs frequently enter communication shadow areas, where the cellular communication is unstable, we need to perform all these well before it happens (see Section~\ref{sec:nec} and Fig.~\ref{fig:communication}). This prediction is crucial in our business because we are paid by the number of impressions only to advertiser's demographic targets.
    
Fortunately, similar problems have been solved by researchers. Next-Point-of-Interest (Next-POI) predictions are one of the most popular research topics in the field of location-based recommender systems~\cite{10.1145/3132847.3133036,DBLP:journals/corr/ZhangLWSYL17,10.1145/3132847.3133056,8915810,10.1145/3357384.3358171,10.1145/3178876.3186058,ijcai2020-445,9133505}. Many different approaches have been proposed to accomplish the task. They are mostly recurrent neural network (RNN)-based models. Two of the state-of-the-art models are \emph{DeepMove}~\cite{10.1145/3178876.3186058} and \emph{Flashback}~\cite{yang2020location} whose research goals are the most similar to ours. However, their model efficiencies are not as high as our model in some datasets.

To this end, we propose an efficient model to predict next locations. Since we deliver millions of impressions per month and will deliver tens of millions of impressions per month soon according to a current growth rate, even a small prediction error rate may yield a non-trivial loss. Therefore, we pursue a method balanced between model size and accuracy. Our proposed model consists of four modules as shown in Fig.~\ref{fig:archi}: i) an attentive short-term history module, ii) a long-term history module without attention, iii) a main neural ordinary differential equation (NODE) layer to generate the last hidden vector $\bm{h}(1)$ from the initial hidden vector $\bm{h}(0)$, and iv) a classifier to predict next locations.

Let $t_{K}$ be the current time in a day, when we want to predict the next $M$ locations of a taxicab. In the first module, we process the taxicab's recent $K$ locations at $\{t_1, \cdots, t_{K}\}$ with a dot-product-based attention mechanism --- we call this set of $K$ recent locations as \emph{session}. In our work, both the session length $K$ and the number of future locations to predict $M$ can be varied, which make our method more practical. A similar thing happens in the second module with all other past locations, but we do not use attention here because it leads to better efficiency. Each module produces an initial vector from the short and long-term history, respectively. These two initial vectors are concatenated, denoted $\bm{h}(0)$ in the figure, and processed by a NODE layer to generate another hidden vector $\bm{h}(1)$, which will be concatenated with the taxicab embedding. Then, the last hidden vector is processed by the fourth module and we finally predict $M$ next locations simultaneously. The most demanding part is how to design the NODE module because it processes all information. In fact, it occupies the biggest portion in our design.

The reason why we adopt NODEs to process the initial vector and derive the last hidden vector is twofold. First, we consider a limited set of information to predict next locations. In general, next locations are partially influenced by auxiliary environments, e.g., weather conditions, etc. To make our method applicable to any environments, however, we do not consider such auxiliary information while predicting next locations. To accomplish the challenging task with such limited information, we adopt NODEs which learn smooth homeomorphic functions (see Section~\ref{sec:node}) and are well-posed under the mild condition of locally Lipschitz (see Section~\ref{sec:train}). A similar design philosophy had been adopted in the field of recommender systems. For instance, LightGCN~\cite{10.1145/3397271.3401063} uses only linear layers to process user/item embeddings and outperforms other non-linear collaborative filtering (CF) models because of the information-scarce environments of the task. According to them, highly non-linear models with overcapacity are easily over-fitted to data and hard to regularize in the field of CF. Our problem can also be understood as a geographical recommendation problem between locations and taxicabs because the relationship between them is analogous to that between items and users in CF. Therefore, we conjecture that their design philosophy holds in our setting and adopt NODEs to design our model. Our experimental results well justify our conjecture.

Second, NODEs are known for its good trade-off between model accuracy and size. In other fields, including computer vision, time-series forecasting, and so on, there have been reported multiple times that NODE-based models outperform existing conventional models and/or have a much higher model efficiency than them~\cite{2019arXiv191010470P,debrouwer2019gruodebayes}. We also aim at designing a lightweight model (in terms of the inference time) that shows a reasonable model accuracy.

However, it is still a hard problem how to design a lightweight NODE model with a sufficient model accuracy. In the basic forms of NODEs, we use a fixed set of parameters, denoted $\bm{\theta}_{fixed}$. In our advanced model, however, the parameters of NODEs are dynamically generated from input, denoted $\bm{\theta}_{adaptive} = g(\bm{\theta}_{fixed}, \bm{h}(0); \bm{\theta}_{g})$ where $g$ is a parameter generation network and $\bm{h}(0)$ is an initial hidden vector. Note that $g$ is to create a variation of $\bm{\theta}_{fixed}$ for $\bm{h}(0)$ rather than generating from scratch. In this way, we can generate a reliable set of parameters.

We conduct experiments with three datasets. Two of them are popular benchmark datasets for next-POI recommendation and one is our taxicab dataset, collected Fall 2020 with 177 taxicabs in the largest metropolitan area of South Korea. Our method shows the best accuracy in all datasets with much smaller model costs. Our contributions are as follows:
\begin{enumerate}
    \item We design a lightweight yet accurate next-POI recommendation model based on NODEs.
    \item We design our own NODE layer with various techniques: i) continuous \texttt{GRU} cells, ii) ODE state jumps, and iii) adaptive parameter generation.
    \item Our model shows a near perfect accuracy for our dataset.
\end{enumerate}

\section{Preliminaries \& Related Work}
We introduce our literature survey and preliminary knowledge to understand our work.

\subsection{Neural Ordinary Differential Equations (NODEs)}\label{sec:node}
NODEs solve the following integral problem to calculate $\bm{h}(t_{i+1})$ from $\bm{h}(t_i)$~\cite{NIPS2018_7892}:
\begin{linenomath*}\begin{align}
    \bm{h}(t_{i+1}) = \bm{h}(t_i) + \int_{t_i}^{t_{i+1}}f(\bm{h}(t),t;\bm{\theta}_f)dt,
\end{align}\end{linenomath*}where $f(\bm{h}(t),t;\bm{\theta}_f)$, which we call \emph{ODE function}, is a neural network to approximate $\dot{\bm{h}} \stackrel{\text{def}}{=} \frac{d \bm{h}(t)}{d t}$. To solve the integral problem, NODEs rely on ODE solvers, e.g., the explicit Euler method, the Dormand--Prince (DOPRI) method, and so forth~\cite{DORMAND198019}.

Therefore, NODEs are typically understood as a continuous generation of neural networks. In conventional neural networks, the time variable $t$ for $\bm{h}(t)$ is typically an non-negative integer. In NODEs, however, $t$ can be any arbitrary real number.

Let $\phi_t : \mathbb{R}^{\dim(\bm{h}(t_0))} \rightarrow \mathbb{R}^{\dim(\bm{h}(t_m))}$ be a mapping from $t_0$ to $t_m$ created by an ODE after solving the integral problem. It is well-known that $\phi_t$ becomes a homeomorphic mapping: $\phi_t$ is continuous and bijective and $\phi_t^{-1}$ is also continuous for all $t \in [0,T]$, where $T$ is the last time point of the time domain~\cite{NIPS2019_8577,massaroli2020dissecting}. From this characteristic, it is known that the topology of the input space of $\phi_t$ is preserved in its output space, and therefore, trajectories crossing each other cannot be represented by NODEs.



While preserving the topology, NODEs can perform machine learning tasks and it was shown in~\cite{yan2020robustness} that it increases the robustness of representation learning to adversarial attacks and unexpected inputs. We consider that this characteristic is also suitable for learning reliable hidden representations from past locations, when there is no abundant information. As mentioned earlier, our setting includes only taxicab locations without auxiliary information such as weather conditions, local events, and so on. We conjecture that NODEs that learn smooth homeomorphic functions are suitable for our challenging task as in~\cite{10.1145/3397271.3401063,choi2021ltocf,kim2021oct,jhin2021acenode}.


Instead of the backpropagation method, the adjoint sensitivity method is used to train NODEs for its efficiency and theoretical correctness~\cite{NIPS2018_7892}. After letting $\bm{a}_{\bm{h}}(t) = \frac{d L}{d \bm{h}(t)}$ for a task-specific loss $L$, it calculates the gradient of loss w.r.t model parameters with another reverse-mode integral as follows:\begin{align*}\nabla_{\bm{\theta}_f} L = \frac{d L}{d \bm{\theta}_f} = -\int_{t_m}^{t_0} \bm{a}_{\bm{h}}(t)^{\mathtt{T}} \frac{\partial f(\bm{h}(t), t;\bm{\theta}_f)}{\partial \bm{\theta}_f} dt.\end{align*}

It is known that NODEs have a couple of advantages. First, NODEs can sometimes significantly reduce the required number of parameters when building neural networks~\cite{2019arXiv191010470P}. Second, NODEs enable us to interpret the time variable $t$ as continuous, which is discrete in conventional neural networks~\cite{NIPS2018_7892}.


\paragraph{Homeomorphism and Model Reliability}
It had been recently reported that the homeomorphic characteristic of NODEs brings non-trivial model reliability since it is guaranteed that NODEs produce similar hidden vectors for two similar inputs~\cite{yan2020robustness,NIPS2019_8577}. In addition, LightGCN~\cite{10.1145/3397271.3401063} was successful for collaborating filtering by adopting simple neural network architectures to prevent overfitting for sparse recommendation datasets. Most next POI recommendation datasets are also sparse since they contain only past trajectories without any auxiliary information. Inspired by LightGCN, we adopt NODEs to design our main module to process hidden vectors to prevent the situation that two similar trajectory patterns are mapped to too much different last hidden vectors.



\subsection{ODE Solvers} There is an analogy between residual/dense connections and ODE solvers. ODE solvers discretize time variable $t$ and convert an integral into many steps of additions~\cite{10.2307/j.ctvzsmfgn}. For instance, the explicit Euler method can be written as follows in a step:
\begin{linenomath*}\begin{align}\label{eq:euler}
\bm{h}(t + s) = \bm{h}(t) + s \cdot f(\bm{h}(t), t;\bm{\theta}_f),
\end{align}\end{linenomath*}where $s$, which is usually smaller than 1, is a configured step size of the Euler method. Note that this equation is identical to a residual connection when $s=1$. To derive $\bm{h}(1)$ from $\bm{h}(0)$ with $s=\frac{1}{2}$, for instance, we need two steps with the explicit Euler method.

Other ODE solvers use more complicated methods to update $\bm{h}(t + s)$ from $\bm{h}(t)$. For instance, the fourth-order Runge--Kutta (RK4) method uses the following method:
\begin{linenomath*}\begin{align}\label{eq:rk4}
\bm{h}(t + s) = \bm{h}(t) + \frac{s}{6}\Big(f_1 + 2f_2 + 2f_3 + f_4\Big),
\end{align}\end{linenomath*}where $f_1 = f(\bm{h}(t), t;\bm{\theta}_f)$, $f_2 = f(\bm{h}(t) + \frac{s}{2}f_1, t+\frac{s}{2};\bm{\theta}_f)$, $f_3 = f(\bm{h}(t) + \frac{s}{2}f_2, t+\frac{s}{2};\bm{\theta}_f)$, and $f_4 = f(\bm{h}(t)+sf_3, t+s;\bm{\theta}_f)$.

It is also known that dense convolutional networks (DenseNets~\cite{zhu2019convolutional}) and fractal neural networks (FractalNet~\cite{Larsson2017FractalNetUN}) are similar to RK4 (as so are residual networks to the explicit Euler method)~\cite{pmlr-v80-lu18d}.



\subsection{Next-POI Recommendation}

Next-POI recommendation is a popular topic and has many applications. Among them, the next-POI recommendation attracted much attention from the community and much research work has been conducted.


Zhao et al. designed a tensor factorization-based method with user-POI, POI-time, and POI-POI interaction records, which is one of the early work in this field~\cite{10.5555/3015812.3015858}. He et al. also designed a similar tensor-based latent model to grasp latent check-in patterns and predict future patterns~\cite{10.5555/3015812.3015833}. Feng et al. designed a special embedding-based mechanism for POI recommendation~\cite{10.5555/2832415.2832536}. Liu et al. used a recurrent neural network (RNN) to capture spatial and temporal information~\cite{10.5555/3015812.3015841}. Kong et al. and Zhao et al. integrated temporal and spatial gating mechanism into a RNN-based architecture~\cite{10.5555/3304889.3304985,9133505}. Chang et al. also proposed an advanced embedding method for POI recommendation~\cite{10.5555/3304222.3304226}. Wang et al. developed a POI recommendation algorithm for mobile devices~\cite{10.1145/3366423.3380170}. Manotumruksa et al. proposed a factorization-based method for collaborative filtering but their method shows good performance for next-POI recommendation~\cite{10.1145/3132847.3133036}. However, all these methods do not explicitly consider attention among check-in patterns. In order to consider dependencies, Feng et al. proposed \emph{DeepMove} which has a special attention mechanism~\cite{10.1145/3178876.3186058}. Flashback~\cite{yang2020location} is one of the most recent methods, which shows the state-of-the-art accuracy in some cases.

\section{Taxicab Rooftop Advertising}\label{sec:service}

\begin{figure}[t]
	\centering
	\includegraphics[width=1\columnwidth]{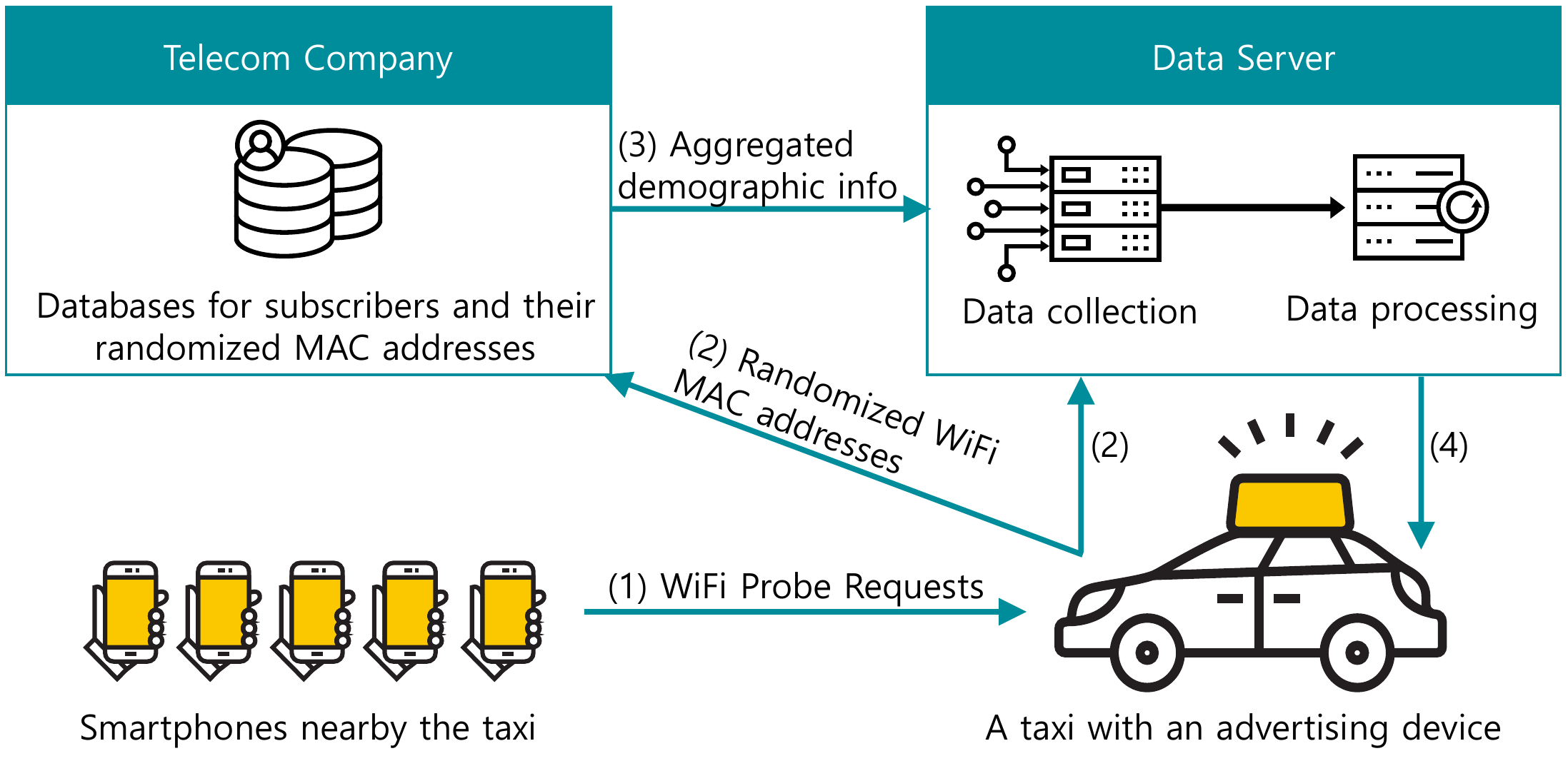}
	\caption{The service workflow of our taxicab rooftop advertising. See Section~\ref{sec:service} for its detailed descriptions.}
	\label{fig:data_curation}
\end{figure}

Motov is a location-based advertisement start-up in South Korea. This company works with one of the largest telecommunication companies in South Korea to detect real-time demographic information around taxicabs.
Taxicabs use this demographic information to provide targeted advertisements that are streamed through the rooftop displays. Overall, there are a thousand of taxicabs deployed in the largest metropolitan area of South Korea. 

Each taxicab has a WiFi intelligent access point (IAP), a Long-term Evolution (LTE) cellular modem, and a GPS.
IAP is a customized version of the general WiFi access point (AP) that captures MAC addresses of WiFi devices such as smartphones in real-time. Smartphones send a probing message periodically to find nearby WiFi access points. The rooftop device with a LCD display on a taxicab detects WiFi probe messages. The transmission delay of the probing message is under 50 microseconds in general, and the signal traverses through the air in the speed of light~\citep{10.1007/s11276-009-0192-z}. The signal decoding time in the collecting device is also under a few microseconds. Therefore, a probing message is received instantaneously, and data is curated in real-time, even at a taxicab velocity of 100km/h. The taxicab reports discovered MAC addresses to the biggest telecommunication company in South Korea, which will respond to us with the anonymized demographic statistics of age and gender for the MAC address owners.

To protect user privacy during the data acquisition process, the demographic information is collected in an aggregated manner. We do not collect each individual demographic information. The detailed service process is as follows. The numbers for the following steps correspond to those in the workflow in Fig.~\ref{fig:data_curation}. 
(1) Individual smartphones keep sending WiFi probe requests to find a new AP. 
Because the IAP on a taxicab keeps scanning these WiFi probe request messages, it can extract an MAC address from a request message. 
(2) The taxicab sends the extracted MAC address to both the telecommunication company and the data server of the start-up company. 
(3) The telecommunication company looks up its subscriber database, as MAC addresses are received, and aggregates the demographic information (i.e., age and gender) every minute. 
Then, they send the aggregated demographic data to the start-up's server every minute. 
As a result, the returned demographic data does not include any original MAC addresses. 
The start-up company ensures privacy by making it impossible to join taxicabs' sensory data and the returned demographic data. 
Finally, (4) the returned demographic data is used for targeted advertisements that are streamed through the rooftop display. We also measure the number of successful exposure to target audiences with this data.


\begin{figure}[t]
	\centering
	\includegraphics[width=1\columnwidth]{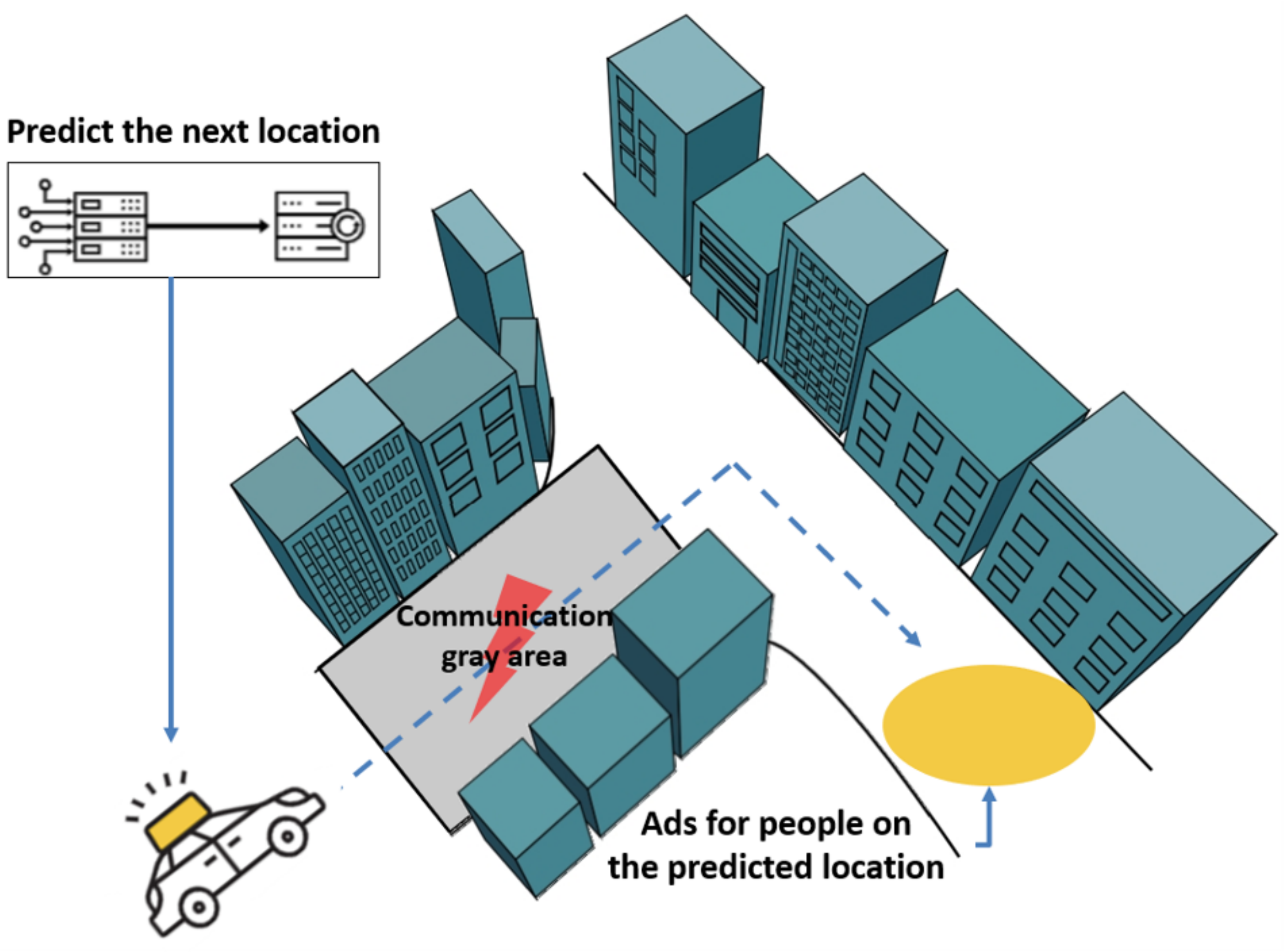}
	\caption{The necessity of NextPoI Recommendation in Rooftop Advertising. See Section~\ref{sec:nec} for its detailed descriptions.}
	\label{fig:communication}
\end{figure}

\subsection{Necessity of Next POI Recommendation in Rooftop Advertising}\label{sec:nec}
However, it is not guaranteed that we can perform the steps (2-4) in the previous subsection promptly due to communication gray areas. Our main service target is the largest metropolitan area of South Korea, which have many high-rise buildings, and as a result, we suffer from many communication gray areas as shown in Fig.~\ref{fig:communication}. In addition, it costs much expense to transfer advertising videos to taxicabs via a high-speed cellular network. Therefore, we rely on a relatively cheaper usage-based cellular network. For these reasons, it is practically impossible to implement real-time advertising services and we provide pseudo-real-time services. Therefore, it is crucial to predict where each taxicab will be shortly and make it ready with the best advertisement whose demographic targets are the best matched with its predicted location. Since each advertisement is paid by the number of impressions to his/her demographic targets, this task is crucial in our business. Therefore, our utmost interest is predicting taxicab locations in several minutes, i.e., five minutes in our experiments.

The other urgent prediction is to predict the current demographic information for each location because it is not the case that each location is frequently visited by our taxicabs. However, this is out of the scope of this paper.

\section{Problem Definition}

Let $C$ be a set of taxicabs. Let $x_t \in X$ be a location (more precisely, one-hot vector denoting a POI location) of a taxicab $c \in C$ at time $t$, where $X$ is a set of all possible POIs. Given a set of recent $K$ locations $\{x_{t_1}, \cdots, x_{t_K}\}$ for $c$, which we call \emph{session} or \emph{short-term history}, and its all other long-term past locations before the recent session, our task is to predict the next $M \leq K$ locations that will be visited by the taxicab $c$. Due to the challenging nature of the problem, we do not consider the case where $M > K$.

We consider the following two cases of time: i) a pair of $t_i$ and $t_{i+1}$ has a fixed time interval for all $i$, or ii) the time interval is varying from one pair to another. Our taxicab application falls into the first case. However, the second case also happens frequently in real-world applications. Thus, we consider both cases in our work to increase the applicability of our method.

One more requirement in our design is, while not sacrificing the model accuracy\footnote{One percentage accuracy drop may lead to a non-trivial loss in our annual revenue.}, to reduce the cost of model in terms of the inference time. Since our platform will run five thousands of taxicabs by the end of this year and tens of thousands by the end of next year, the prediction model should be lightweight in terms of the above model cost definitions.

\begin{definition}[Session]
A session means a set of locations (with their timestamps) that are visited for a single purpose. In our taxicab application, we simply use recent $K$ locations to approximate a session. In the variable time-interval applications, a new session begins if the recent interval is larger than a threshold, which is a widely-used heuristic~\cite{10.1145/3132847.3133056,10.1145/3178876.3186058}.
\end{definition}

\begin{definition}[Trajectory]
Given a taxicab $c \in C$, a trajectory $\mathcal{T}_c$ means a sequence of sessions created by the taxicab. One taxicab generates one long trajectory in our dataset.
\end{definition}

\section{Proposed Method}
In this section, we introduce our model, called \emph{LightMove}. We first describe the overall architecture and then each part in detail.

\subsection{Overall Architecture}
Suppose a taxicab and its past locations. We want to predict its future locations. To this end, we design the model in Fig.~\ref{fig:archi}. The several key design points

First, we learn an embedding vector for each location $x \in X$. These embedding vectors are fed into our model given an input set of locations. We also learn an embedding vector for each taxicab.

Second, we separate the short-term and the long-term history while processing. The short-term location information, i.e., a recent session, is processed by a dot-product-based attention, and the same processing happens for the remaining long-term location information without attention. We concatenate them into $\bm{h}(0)$.

With the initial ODE state $\bm{h}(0)$, third, we use a NODE-based module to derive $\bm{h}(1)$. This module is the main processing area in our model and plays a key role in our task. We intentionally choose a NODE-based design at this step for its appropriateness to our task in that i) it learns a smooth homeomorphic mapping which is suitable for our information-scarce environment, and ii) it shows a low model cost.

Fourth, $\bm{h}(1)$ is concatenated with a taxicab embedding vector to constitute the last hidden vector, which will be used for our next $M$ location predictions.

\subsection{Attentive History Processing Modules}
In this subsection, we describe attentive history processing modules for short and long-term past location information.

Let $S = \{x_{t_1}, \cdots, x_{t_K}\}$ be a recent session of a taxicab $c \in C$ --- $x_{t_i}$ is typically a one-hot vector denoting a POI with a timestamp, which constitute a short-term history. Our first module is to properly process this session information to extract an initial short-term vector. We use the following dot-product-based attention:
\begin{align}
    \bm{E}_S &= \texttt{lookup}(S),\\
    \bm{A}_S &= \sigma(\bm{E}_S \cdot \bm{E}_S^\intercal),
\end{align}where \texttt{lookup} is a location/time embedding lookup function, which outputs an embedding matrix $\bm{E}_S \in \mathbb{R}^{|S| \times (D_{loc}+D_{time})}$ with a drop-out rate $\gamma$. $D_{loc},D_{time}$ mean the dimensionality of embedding for location and time --- we divide the 24h-hour band into slots and learn an embedding vector for each slot. In other words, $\bm{E}_S$ contains all those embedding vectors for the locations in $S$. $\sigma$ is a softmax function. Therefore, $\bm{A}_S \in [0,1]^{|S| \times |S|}$ is a matrix which contains pair-wise attention values among all those short-term locations. After that, we use the following operation to derive an initial short-term matrix:
\begin{align}
    \bm{H}_S = \bm{A}_S\bm{E}_S,
\end{align}where $\bm{H}_S \in \mathbb{R}^{|S| \times (D_{loc}+D_{time})} $ is a matrix, each row of which is an initial vector for each location in $S$.

We perform the same thing for a long-term location information $L$ and derive an initial long-term matrix $\bm{H}_L$. However, we do not use any attention here because it increases the cost significantly and makes the entire model hard to train. In general, $|L| > |S|$ and it requires much computation to process long-term history. Therefore, we strategically drop the attention for the long-term history processing. The initial vector $\bm{H}_{init} \in \mathbb{R}^{(|S| + |L|) \times (D_{loc}+D_{time})}$ is a concatenation of them as follows:
\begin{align}
    \bm{H}_{init} = \bm{H}_S \oplus \bm{H}_L,
\end{align} where $\oplus$ is a concatenation operation.

Among various other possible attention alternatives, we use the dot-product-based pair-wise attention mechanism for its appropriateness to recommendation. In one of the most recent collaborative filtering research directions, for instance, they consider that a set of items purchased by a user creates a clique of items via the user, whose embedding vectors are highly correlated~\cite{10.1145/3331184.3331267,Chen_Wu_Hong_Zhang_Wang_2020,10.1145/3397271.3401063}. They explicitly build a bipartite graph between user and items and utilize its normalized Laplacian matrix to aggregate neighbors' embedding vectors. Our proposed concept is somehow aligned with the notion but use a trainable attention method rather than relying on a fixed graph architecture.

\subsection{NODE \& Classification Modules}
In the previous subsection, we showed how to calculate the initial matrix, denoted $\bm{H}_{init}$. In this subsection, we show how to evolve each initial vector (i.e., each row of $\bm{H}_{init}$) with a NODE-based layer. With the $i$-th initial vector (row) $\bm{h}_i$ of $\bm{H}_{init}$, we use a NODE-based layer to derive its hidden vector as follows:
\begin{align}
    \bm{j}_i(t_1) &= \bm{h}_i(0) + \int_0^{t_1} f(\bm{h}_i(t),t,\bm{\theta}_{fixed}) dt,\\
    \bm{h}_i(t_1) &= \texttt{GRU}(\bm{j}_i(t_1); \bm{\theta}_{\texttt{GRU}}),\label{eq:jump1} \\ 
    \bm{j}_i(t_2) &= \bm{h}_i(t_1) + \int_{t_1}^{t_2} f(\bm{h}_i(t),t,\bm{\theta}_{fixed}) dt,\\
    \bm{h}_i(t_2) &= \texttt{GRU}(\bm{j}_i(t_2); \bm{\theta}_{\texttt{GRU}}),\label{eq:jump2}\\
    \vdots\\
    \bm{j}_i(1) &= \bm{h}_i(t_{J}) + \int_{t_{J}}^{1} f(\bm{h}_i(t),t,\bm{\theta}_{fixed}) dt,\\
    \bm{h}_i(1) &= \texttt{GRU}(\bm{j}_i(1); \bm{\theta}_{\texttt{GRU}}),\label{eq:jump3}
\end{align} where $\bm{h}_i(0) = \bm{h}_i$, $0 < t_i < t_j < 1$, where $i < j$, are intermediate time points, and $J$ is the number of jumps between $t=0$ and $t=1$. In other words, our NODE layer evolves each initial vector (row) of $\bm{H}_{init}$ to derive its hidden vector with multiple jumps. Each jump is done by \texttt{GRU} in our design. The jump mechanism is frequently used in conjunction with NODEs~\cite{NIPS2019_8773,debrouwer2019gruodebayes}. Our design for the ODE function $f$, which approximates $\frac{d \bm{h}_i(t)}{d t}$, is as follows:
\begin{align}
    \bm{r}_i(t) &= \Phi(\bm{W}_r \bm{h}_i(t) + \bm{U}_r \bm{h}_i(t) + \bm{b}_r), \label{eq:gru1} \\ 
    \bm{z}_i(t) &= \Phi(\bm{W}_z \bm{h}_i(t) + \bm{U}_z \bm{h}_i(t) + \bm{b}_z), \label{eq:gruz}\\
    \bm{m}_i(t) &=  \texttt{tanh}(\bm{W}_m \bm{h}_i(t) + \bm{U}_m (\bm{r}_i(t) \odot \bm{h}_i(t)) + \bm{b}_m),\\
    \frac{d \bm{h}_i(t)}{d t} &=  (1 - \bm{z}_i(t)) \odot ( \bm{m}_i(t) - \bm{h}_i(t)), \label{eq:gru2}
\end{align}where $\odot$ is an elementwise multiplication, $\Phi$ is a sigmoid activation. All those $\bm{W},\bm{U},\bm{b}$ parameters are trainable and denoted as $\bm{\theta}_{fixed}$. Our specific choices in Eqs.~\eqref{eq:gru1} to~\eqref{eq:gru2} are a continuous generalization of gated recurrent units (GRUs)~\cite{debrouwer2019gruodebayes}.

Let $\bm{H}$ be the derived hidden matrix, the $i$-th row of which corresponds to $\bm{h}_i(1)$. We use the following classifier to predict the future $M$ locations that will be visited by the target taxicab $c$:
\begin{align}
    \bm{H}_{last} &= \phi(\bm{H} \oplus \bm{e}_{c}),\\
    \bm{P} &=  \sigma(\texttt{FC}_{last}(\bm{H}_{last}; \bm{\theta}_{last})),
\end{align}where $\phi:\mathbb{R}^{(|S|+|L|)\times(D_{loc}+D_{time} + D_{taxi})} \rightarrow \mathbb{R}^{M \times(D_{loc}+D_{time} + D_{taxi})}$ is a resize operation, $\bm{e}_c$ is the embedding of the taxicab $c$, and $\texttt{FC}_{last}: \mathbb{R}^{M \times(D_{loc}+D_{time} + D_{taxi})} \rightarrow \mathbb{R}^{M \times |X|}$ is a fully connected layer followed by a softmax. $\bm{P} \in [0,1]^{M \times |X|}$ contains future $M$ location predictions for the taxicab $c$. Each row of $\bm{P}$ corresponds to a standard classification over $X$. The resize operator $\phi$ can be designed as a fully connected layer, a matrix slice operation, and so forth.

\subsection{Adaptive Parameter Generation}
Our above design outperforms many baseline methods in our experiments. To further increase the model accuracy, we design an advanced architecture with generating parameters as follows:
\begin{align}
    \bm{\theta}_{adaptive} &= g(\bm{\theta}_{fixed},\bm{h}_i(0); \bm{\theta}_{g}),\\
    \bm{h}_i(t_{j+1}) &= \bm{h}_i(t_j) + \int_{t_j}^{t_{j+1}} f(\bm{h}_i(t),t,\bm{\theta}_{adaptive})dt, \textrm{ for each $j$ },
\end{align}where $g:\mathbb{R}^{\dim(\bm{\theta}_{fixed})} \rightarrow \mathbb{R}^{\dim(\bm{\theta}_{adaptive})}$ is a parameter generation function. $\dim(\bm{\theta}_{fixed}) = \dim(\bm{\theta}_{adaptive})$ in this function.

This adaptive parameter generation greatly increases the model accuracy, when the task is challenging, by producing parameters suitable for a certain input $\bm{h}_i(0)$ in the NODE layer. However, we do not allow a large difference between $\bm{\theta}_{fixed}$ and $\bm{\theta}_{adaptive}$ to make it as close to the original homeomorphic mapping as possible. As mentioned earlier, our NODE layer only with $\bm{\theta}_{fixed}$ learns a homeomorphic function and it can learn reliable representations when the information in input is scarce or contains unexpected patterns. Therefore, we want to stick to the original setting as much as possible even with the adaptive parameter generation. To this end, we generate parameters for $\bm{z}_i(t)$ of Eq.\eqref{eq:gruz} only:
\begin{align}
    \bm{W}'_{z} &= \texttt{resize}(\texttt{FC}_{\bm{W}}(\texttt{vec}(\bm{W}_{z}) \oplus \bm{h}_i(0); \bm{\theta}_{\bm{W}})),\\
    \bm{U}'_{z} &= \texttt{resize}(\texttt{FC}_{\bm{U}}(\texttt{vec}(\bm{U}_{z}) \oplus \bm{h}_i(0); \bm{\theta}_{\bm{U}})),\\
    \bm{b}'_{z} &= \texttt{FC}_{\bm{b}}(\bm{b}_{z} \oplus \bm{h}_i(0); \bm{\theta}_{\bm{b}}),\\
    \bm{z}_i(t) &=  \Phi(\bm{W}'_z \bm{h}_i(t) + \bm{U}'_z \bm{h}_i(t) + \bm{b}'_z),
\end{align}where $\texttt{vec}$ is a vectorization function and $\texttt{resize}$ is a resize function to recover into the original matrix form. $\bm{z}_i(t)$ is called \emph{gating} and plays a key role. Therefore, fine-tuning it for each input is a sensible approach to increase the capacity of the NODE layer if a dataset to predict is complicated, e.g. Foursquare in our experiments.

\subsection{Training Algorithm}\label{sec:train}
The overall training method in Algorithm~\ref{alg:trainalg} is similar to what used in DeepMove~\cite{10.1145/3178876.3186058}. Each taxicab's complete trajectory is divided into a set of training/validating/testing trajectories with a ratio of 70/15/15 in chronological order. A mini-batch is a single training trajectory for a taxicab (user). Due to the large sequence of the long-term history, denoted $L$ in line~\ref{alg:long}, both DeepMove and our LightMove use one training trajectory in an iteration.

However, SERM~\cite{10.1145/3132847.3133056} uses a mini-batch of multiple taxicabs (users) since it does not process the long-term history. It utilizes only the recent session to predict next $M$ locations. However, this strategy shows relatively inaccurate predictions for our taxi dataset. Therefore, we stick to the proposed training method. While training in line~\ref{alg:train}, we use a cross-entropy loss with a $L^2$ regularizer for $\bm{\theta}$.

\begin{algorithm}[t]
\small
\SetAlgoLined
\caption{How to train LightMove}\label{alg:trainalg}
\KwIn{Taxicabs $C$, Training Trajectories $\{\mathcal{T}_c\}_{1}^{|C|}$, Validating Trajectories $\{\mathcal{V}_c\}_{1}^{|C|}$, Initial Learning Rate $\lambda$, Learning Rate Decay Factor $\alpha$, $L^2$ Regularization Coefficient $\beta$}
Let $\bm{\theta}$ be all parameters to learn and Initialize $\bm{\theta}$

\While {the POI classification loss $L$ is not converged}{
\For{Each taxicab $c \in C$}{
    $X$ gets the first 70\% sessions of a training trajectory $\mathcal{T}_c$
    
    $Y$ gets the last 30\% sessions of a training trajectory $\mathcal{T}_c$

    $S \gets$ the recent session of $X$
    
    $L \gets$ all sessions of $X$ before $S$ \label{alg:long}
    
    \tcc{Given $S$ and $L$, our training task is to predict $Y$.}
    Train $\bm{\theta}$ with the short-term history $S$ and the long-term history $L$ to predict $Y$. \label{alg:train}
}
Validate with $\mathcal{V}_c$ of all $c \in C$ to update the best model checkpoint.
}
\Return $\bm{\theta}$;
\end{algorithm}

\paragraph{On the Tractability of Training the NODE layer.} The ODE version of the Cauchy--Kowalevski theorem states that, given $f = \frac{d \bm{h}(t)}{d t}$, there exists a unique solution of $\bm{h}$ if $f$ is analytic (or locally Lipschitz continuous). In other words, the ODE problem is well-posed if $f$ is analytic~\cite{10.2307/j.ctvzsmfgn}. In our case, the function $f$ in Eqs.~\eqref{eq:gru1} to~\eqref{eq:gru2} uses only matrix multiplications and hyperbolic tangent/sigmoid activations that are all analytic. This implies that there will be only a unique optimal ODE, given an initial vector $\bm{h}(0)$. Because of i) the uniqueness of the solution and ii) our analytic definitions of $f$, we believe that our training method can find a good solution.

\section{Experimental Evaluations}
In this section, we introduce our experimental environments and results. All experiments were conducted in the following software and hardware environments: \textsc{Ubuntu} 18.04 LTS, \textsc{Python} 3.6.6, \textsc{Numpy} 1.18.5, \textsc{Scipy} 1.5, \textsc{Matplotlib} 3.3.1, \textsc{PyTorch} 1.2.0, \textsc{CUDA} 10.0, and \textsc{NVIDIA} Driver 417.22, i9 CPU, and \textsc{NVIDIA RTX Titan}.

\subsection{Experimental Environments}
\subsubsection{Datasets}
We use the following three datasets for our experiments. For thorough evaluations, we use not only our taxicab data but also other standard benchmark evaluation data for next-POI recommendation. Their statistics are summarized in Table~\ref{tbl:data}.
\begin{enumerate}
    \item Foursquare contains user check-in data for a year from Feb. 2010 in New York. Every check-in has a user ID, timestamp, and POI ID. We use the pre-processed data in~\cite{10.1145/3178876.3186058}, where low-frequency users are removed and check-in records are divided into sessions. There is at least 72-hour gap between two consecutive sessions after the pre-processing. The average session length is 8.64 check-ins in this datasets. This dataset is one of the most difficult ones for next-POI recommendation because POIs are located closed to each other in New York. Sometimes multiple POIs are in a building.
    \item LA is a Twitter dataset collected in Los Angeles from Aug. to Nov. 2014. It also contains many check-in records after discretizing the city of Los Angeles into many 500m$\times$500m grid cells (POIs). We use the data pre-processed in~\cite{10.1145/3132847.3133056} where two consecutive sessions have at least 10-hour time difference. The average session length is 9.1 check-ins in this dataset.
    \item Taxi is our data collected in the largest metropolitan area of South Korea with 177 taxicabs for a month Fall 2020. Our 177 taxicabs provides 24/7 GPS logs every five minute to our server. One session duration is 45 minutes. Considering missing logs, there are 8.38 logs in a session in average. There are 584 locations that we are interested in. Each location has a size of approximately 100m$\times$100m.
\end{enumerate}

\begin{table}[t]
\centering
\small
\setlength{\tabcolsep}{4pt}
\caption{Statistics of datasets}\label{tbl:data}
\begin{tabular}{cccc}
\specialrule{1pt}{1pt}{1pt}
Name & \#Taxicabs (Users) & \#Locations (POIs) & \#Logs\\ \specialrule{1pt}{1pt}{1pt}
Taxi & 177 & 584 & 353,419 \\
Foursquare & 15,639 & 43,380 &  293,559\\
LA & 153,626 & 734,559 & 1,192,572 \\
\specialrule{1pt}{1pt}{1pt}
\end{tabular}
\end{table}

\subsubsection{Baselines}
A variety of method have been developed so far for POI recommendation. Among them, we consider the following models considering their popularity and impacts:
\begin{enumerate}
    \item RNN, LSTM, BiLSTM, and GRU are classical recurrent neural network models which show good performance in dealing with sequential data~\cite{10.1162/neco.1997.9.8.1735,10.1007/978-3-319-56608-5_46,cho-etal-2014-learning}.
    \item DRCF~\cite{10.1145/3132847.3133036} is a matrix factorization-based method for collaborative filtering but this model can be used for next-POI recommendations as well.  This model does not consider long-term history.
    \item SERM~\cite{10.1145/3132847.3133056} is an advanced recurrent neural network model. However, it does not have any attention mechanism and does not consider long-term history. Both SERM and DRCF are fast because they do not consider long-term history.
    \item DeepMove~\cite{10.1145/3178876.3186058} is one of the state-of-the-art models. This model has its own attention mechanism.
    \item Flashback~\cite{yang2020location} is one of the most recent method in the next POI recommendation, which has a strong point in processing past historical trajectories.
    \item LightMove is our proposed method and it has several variations. We use the following notation to denote those variations. `G' or `L' means we jump through either \texttt{GRU} or \texttt{FC} in Eqs.~\eqref{eq:jump1},~\eqref{eq:jump2}, and~\eqref{eq:jump3}. `0',`2', or `5' means the number of jumps. `E' or `R' means the Euler or the RK4 method to solve our integral problems. `F' stands for generating parameters for fine-tuning our model. For instance, therefore, LightMove(G2EF) means that we use two \texttt{GRU}-based jumps with the Euler method and the parameter generation method.
\end{enumerate}

\subsubsection{Hyperparameters}
We consider the following hyperparameter ranges for our model and we use the recommended settings for the baseline models in their original papers and/or their respected GitHub repositories:
\begin{enumerate}
    \item The location embedding size $D_{loc}$ is in \{50, 100, 300, 500, 700\}.
    \item The taxicab (user) embedding size $D_{taxi}$ is in \{20, 40, 60, 80\}.
    \item The time embedding size $D_{time}$ is in \{5, 10, 20, 40\}.
    \item The size of $\bm{h}(t)$ is  in \{50, 100, 300, 500, 700\}.
    \item The drop-out rate $\gamma$ is in \{0.3, 0.5\}.
    \item The initial learning rate $\lambda$ is in \{0.005, 0.05, 0.1\}. The learning rate decay factor $\alpha$ is 0.9. Therefore, we multiply 0.9 to the learning rate every epoch. The minimum learning rate is 0.0005 and we do not decrease below it.
    \item The $L^2$ regularization weight $\beta$ is 0.00001.
    \item We use the Adam optimizer to train.
\end{enumerate}

\subsubsection{Evaluation Methods}
Given a dataset consisting of multiple taxicabs (users), we divide each taxicab's entire trajectory into a split of training/validating/testing (70:15:15) periods in chronological order. It is guaranteed that if a taxicab exists in a testing case, it also exists in a training set in our setting. Another possible scenario is to train with a set of taxicabs and test with other remaining taxicabs. However, this configuration does not happen in our business. Therefore, we split an entire trajectory of a taxicab into training/validating/testing periods.

In addition, we also split other datasets into the same ratio for training/validating/testing in chronological order for our experiments. In the field of next-POI recommendation, researchers have used several benchmark datasets without validation. They mostly reported the best accuracy observed during training, which is impractical considering our business model. Because we split in chronological order, there does not exist any point that can be adjusted in favor of us only except the splitting ratio. Moreover, a ratio of 70:15:15 is a sort of standard.

When we evaluate our method without validation, our model's accuracy is far higher than what we will report in this paper in Foursquare. As mentioned earlier, however, this score is rather meaningless in practice when we deploy a model. Therefore, we split each dataset into the three pieces and all our source codes and datasets are available in {\color{blue}\url{https://github.com/Jinsung-Jeon/LightMove}} so one can easily reproduce.

\begin{figure}[t]
    \centering
    \includegraphics{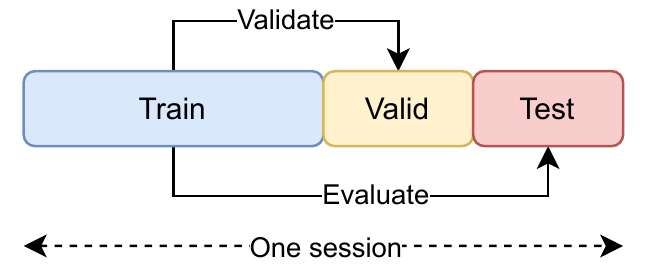}
    \caption{How to validate and test. Given a session of a taxicab, we a session into test/valid/test (70:15:15) periods. }
    \label{fig:test}
\end{figure}

For testing, we further divide a testing trajectory of a taxicab into a set of the first 70\% and a set of the final 30\% locations. Given the first 70\%, we let various models predict the last 30\% locations, which is the same for the training and validating tasks (cf. Fig.~\ref{fig:test}). For evaluation, we use the following accuracy metrics: i) Hits@1/5/10, and ii) mean reciprocal ranking (MRR). Both of them are frequently used to evaluate recommendations. We also report iii) the number of parameters, and iv) the inference time during testing.

\begin{table}[t]
\small
\centering
\setlength{\tabcolsep}{1pt}
\caption{Model performance comparison in Taxi. Since we now record millions of impressions per month and will show tens of millions per month by the end of this year according to the recent growth rate, a few percentage accuracy difference may lead to a large loss.}\label{tbl:acc1}
\begin{tabular}{cccccc}
\specialrule{1pt}{1pt}{1pt}
Method & Hits@1 & Hits@5 & Hits@10 & MRR & \#Params \\ \specialrule{1pt}{1pt}{1pt}
RNN & 0.1872 & 0.4917 & 0.6838 & 0.3411 & 66,154 \\
LSTM & 0.2440 & 0.5769 & 0.7275 & 0.4013 & 96,754\\
BiLSTM & 0.2731 & 0.5918 & 0.7388 & 0.4181 & 192,954\\
GRU & 0.2272 & 0.5303 & 0.7045 & 0.3768 & 86,554\\
SERM & 0.8958 & 0.9988 & \textbf{1.0000} & 0.9422 & 3,815,669\\
DRCF & 0.9581 & \textbf{1.0000} & \textbf{1.0000} & 0.9880 & 248,603\\
DeepMove & 0.9435 & 0.9706 & 0.9759 & 0.9575 & 5,135,534\\
Flashback(RNN) & 0.9688  & 0.9770 & 0.9781 & 0.9729 & 2,262,254\\
Flashback(LSTM) & 0.9913 & 0.9990 & 0.9996 & 0.9946 & 1,271,654 \\
Flashback(GRU) & 0.9934 & 0.9988 & \textbf{1.0000} & 0.9959 & 1,091,054 \\
\hline
LightMove(G0E) & 0.4762 & 0.8611 & 0.9491 & 0.6385 & 1,625,214\\
LightMove(L2E) & 0.5008 & 0.9136 & 0.9757 & 0.6720 & 3,553,704\\
LightMove(G2E) & 0.9985 & \textbf{1.0000} & \textbf{1.0000} & 0.9993 & 3,217,714\\
LightMove(G5E) & 0.9960 & \textbf{1.0000} & \textbf{1.0000} & 0.9979 & 3,321,494\\
\hline
LightMove(G2R) & 0.9983 & \textbf{1.0000} & \textbf{1.0000} & 0.9991 & 3,217,714\\
LightMove(G2EF) & \textbf{0.9988} & \textbf{1.0000} & \textbf{1.0000} & \textbf{0.9994} & 135,868,714\\
\specialrule{1pt}{1pt}{1pt}
\end{tabular}
\end{table}

\begin{table}
\small
\centering
\setlength{\tabcolsep}{1pt}
\caption{Model performance comparison in Foursquare}\label{tbl:acc2}
\begin{tabular}{cccccc}
\specialrule{1pt}{1pt}{1pt}
Method & Hits@1 & Hits@5 & Hits@10 & MRR & \#Params  \\ \specialrule{1pt}{1pt}{1pt}
RNN & 0.0120 & 0.0258 & 0.0408 & 0.0234 & 1,094,334\\
LSTM & 0.0105 & 0.0269 & 0.0418 & 0.0228 & 1,675,643\\
BiLSTM & 0.0082 & 0.0314 & 0.0471 & 0.0217 & 1,519,833\\
GRU & 0.0677 & 0.1406 & 0.2098 & 0.1146 & 1,114,734\\
SERM & 0.0564 & 0.1534 & 0.2098 & 0.1049 & 11,107,796\\
DRCF & 0.0654 & 0.1443 & 0.1781 & 0.1025 & 2,428,603\\
DeepMove & 0.1250 & 0.2553 & 0.2920 & 0.1836 & 20,053,973\\
Flashback(RNN) & 0.1252 & 0.2981 & 0.3548 & 0.2031 & 9,900,593 \\
Flashback(LSTM) & 0.1253 & 0.2887 & 0.3436 & 0.2007 & 10,442,393 \\
Flashback(GRU) &  0.1224 & 0.2938 & 0.3544 & 0.2008 & 10,261,793\\
\hline
LightMove(G0E) & 0.1402 & 0.2435 & 0.2687 & 0.1872 & 6,957,453\\
LightMove(L2E) & 0.1537 & \textbf{0.3203} & 0.3654 & 0.2276 & 2,956,908\\
LightMove(G2E) & 0.1390 & 0.2984 & 0.3523 & 0.2121 & 3,182,573\\ 
LightMove(G5E) & 0.1380 & 0.2994 & 0.3548 & 0.2107 & 20,635,373\\
\hline
LightMove(L2R) & 0.1542 & 0.3187 & \textbf{0.3656} & 0.2257 & 2,956,908\\
LightMove(L2EF) & \textbf{0.1545} & 0.3187 & 0.3653 & \textbf{0.2288} & 4,287,908\\
\specialrule{1pt}{1pt}{1pt}
\end{tabular}
\end{table}

\begin{table}
\small
\centering
\setlength{\tabcolsep}{1pt}
\caption{Model performance comparison in LA}\label{tbl:acc3}
\begin{tabular}{cccccc}
\specialrule{1pt}{1pt}{1pt}
Method & Hits@1 & Hits@5 & Hits@10 & MRR & \#Params  \\ \specialrule{1pt}{1pt}{1pt}
RNN & 0.1357 & 0.2438 & 0.2992 & 0.1841 & 203,817\\
LSTM & 0.0823 & 0.1859 & 0.2239 & 0.1370 & 561,251\\
BiLSTM & 0.1092 & 0.2100 & 0.2609 & 0.1597 & 1,114,881\\
GRU & 0.1625 & 0.2622 & 0.3121 & 0.2179 & 224,217\\
SERM & 0.3173 & 0.4921 & 0.5301 & 0.3938 & 5,732,534\\
DRCF & 0.2243 & 0.3405 & 0.3870 & 0.2803 & 1,733,403\\
DeepMove & 0.3030 & 0.4303 & 0.4662 & 0.3615 &15,743,101\\
Flashback(RNN) & 0.2913 & 0.4421 & 0.4913 & 0.3587 & 7,227,721\\
Flashback(LSTM) & 0.2811 & 0.4336 & 0.4843 & 0.3492 & 7,769,521\\
Flashback(GRU) & 0.2938 & 0.4375 & \textbf{0.4865} & 0.3607 & 7,588,921\\
\hline
LightMove(G0E) & 0.3207 & 0.4290 & 0.4549 & 0.3700 & 984,401\\ 
LightMove(L2E) & 0.3167 & \textbf{0.4431} & 0.4758 & \textbf{0.3756} & 1,953,838\\ 
LightMove(G2E) & 0.3025 & 0.4336 & 0.4698 & 0.3625 & 6,274,811\\
LightMove(G5E) & 0.2954 & 0.4330 & 0.4689 & 0.3574 & 5,878,341\\ 
\hline
LightMove(G0R) & 0.3165 & 0.4246 & 0.4499 & 0.3668 & 984,401\\
LightMove(G0EF) & \textbf{0.3209} & 0.4287 & 0.4536 & 0.3702 & 173,791,333\\
\specialrule{1pt}{1pt}{1pt}
\end{tabular}
\end{table}

\subsection{Experimental Results}

\begin{table}
\small
\centering
\setlength{\tabcolsep}{1pt}
\caption{Inference time comparison}\label{tbl:time1}
\begin{tabular}{ccccccccc}
\specialrule{1pt}{1pt}{1pt}
\multirow{2}{*}{Method} & \multirow{2}{*}{DeepMove} & \multicolumn{3}{c}{Flashback} && \multicolumn{3}{c}{LightMove} \\ 
\cline{3-5}\cline{7-9}
 &  & RNN & LSTM & GRU & &G0E & L2E & G2E \\\specialrule{1pt}{1pt}{1pt}
Taxi & 568s & 349s & 359s & 361s& &112s & 117s& 162s\\
Foursquare & 65s & 119s & 91s & 118s &&17s &21s & 33s \\
LA & 254s & 91s & 118s &91s& & 35s & 39s& 52s\\
\specialrule{1pt}{1pt}{1pt}
\end{tabular}
\end{table}

In Table~\ref{tbl:acc1}, we list the results for Taxi. RNN/LSTM/GRU-based methods all show poor performance. Among baselines, DRCF, DeepMove, and Flashback work well. However, their overall performance is worse than LightMove(G2E), i.e., LightMove with two GRU-based jumps and the Euler ODE solver. LightMove(G2E) has much higher efficiency than LightMove(G2R), considering the simplicity of the Euler method.

Table~\ref{tbl:acc2} summarizes the results in Foursquare. DeepMove and Flashback show the state-of-the-art performance among all baselines. However, LightMove(L2EF) significantly outperforms them with a much smaller number of parameters.

For LA, SERM shows the best performance among all baselines in Table~\ref{tbl:acc3}. LightMove(G0E) shows the best efficiency --- its accuracy is slightly worse than LightMove(G0EF) but has a much smaller number of parameters. From these, we can know that simple models work well for LA. The movement patterns of users in this dataset is not as active as those in other datasets and simple models are enough to predict. Models with high capacity, such as DeepMove, Flashback, and some LightMove variations, are quickly over-fitted to the training data and hard to regularize.

In Table~\ref{tbl:time1}, we compare the runtime of various important methods. In general, our method shows the smallest inference time.

\subsection{Ablation \& Sensitivity Analyses}
\paragraph{\texttt{GRU} vs. \texttt{FC}-based Jumps} Our method provides two options for the jumps in Eqs.~\eqref{eq:jump1},~\eqref{eq:jump2}, and~\eqref{eq:jump3}. The \texttt{GRU}-based jumps show better accuracy than those with \texttt{FC} in many cases. Therefore, our default model uses the \texttt{GRU}-based jumps.

\paragraph{The Number of Jumps} We found that too many jumps sometimes cause over-fitting. In Taxi, for instance, LightMove(G2E) shows higher performance than LightMove(G5E) in all metrics. In Foursquare, 2 jumps also outperform 5 jumps although it has more complicated trajectories than Taxi --- i.e., many POIs are located close to each other in New York and correctly predicting users' next POIs is far more challenging than other cases. Therefore, we recommend 2 jumps in general.

\paragraph{Euler vs. RK4.} According to our results, RK4 and the Euler method show  comparable accuracy, e.g., an MRR of 0.9993 by LightMove(G2E) vs. 0.9991 by LightMove(G2R) in Taxi. Considering the high computational overhead of RK4, we think that the Euler method is a sensible choice in practice.

\begin{figure}
\centering
\subfigure[Hits@1 in Taxi]{\includegraphics[width=0.48\columnwidth,trim={0 0 0 0},clip]{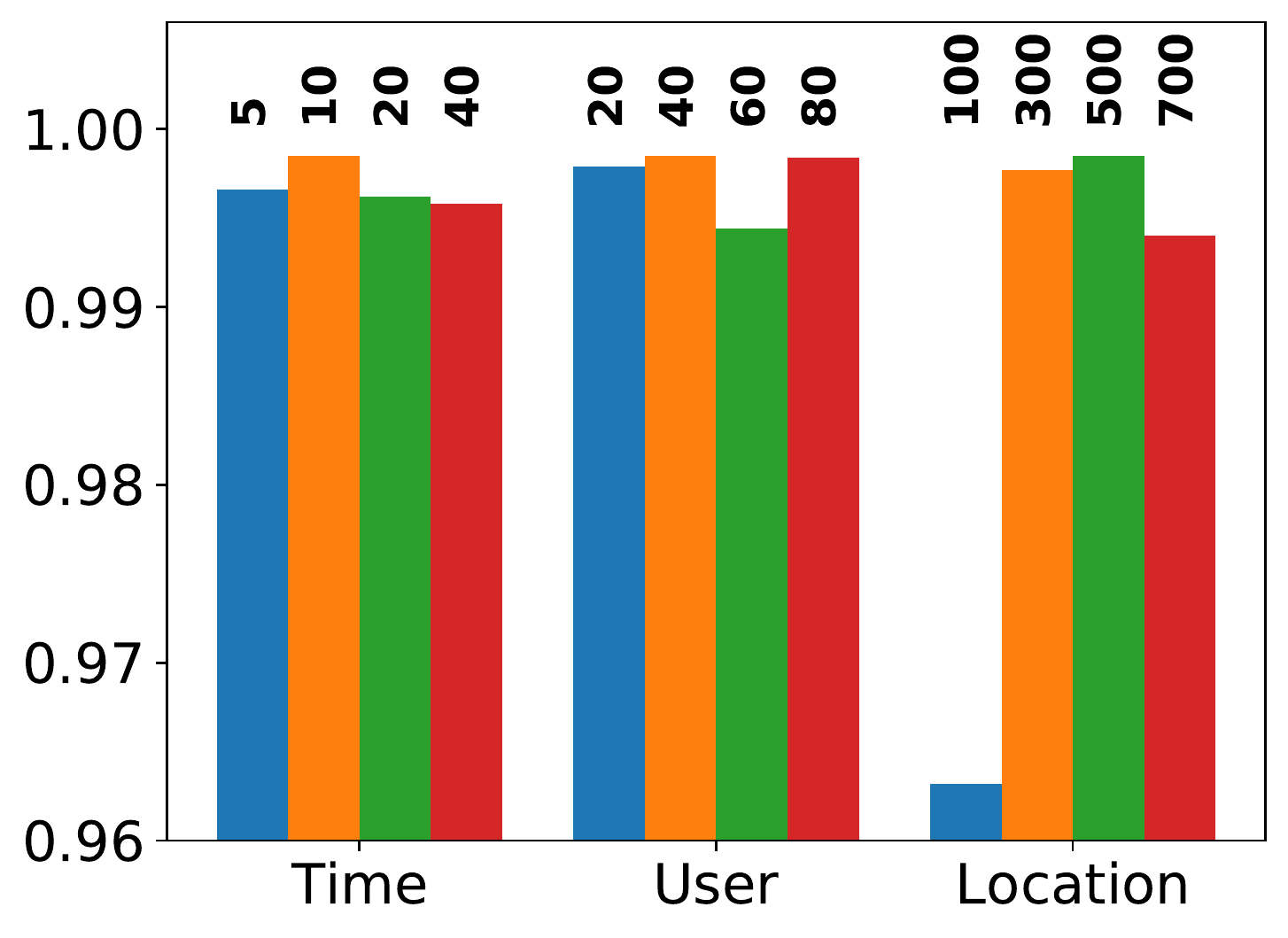}}
\subfigure[Hits@1 in LA]{\includegraphics[width=0.485\columnwidth,trim={0 0 0 0},clip]{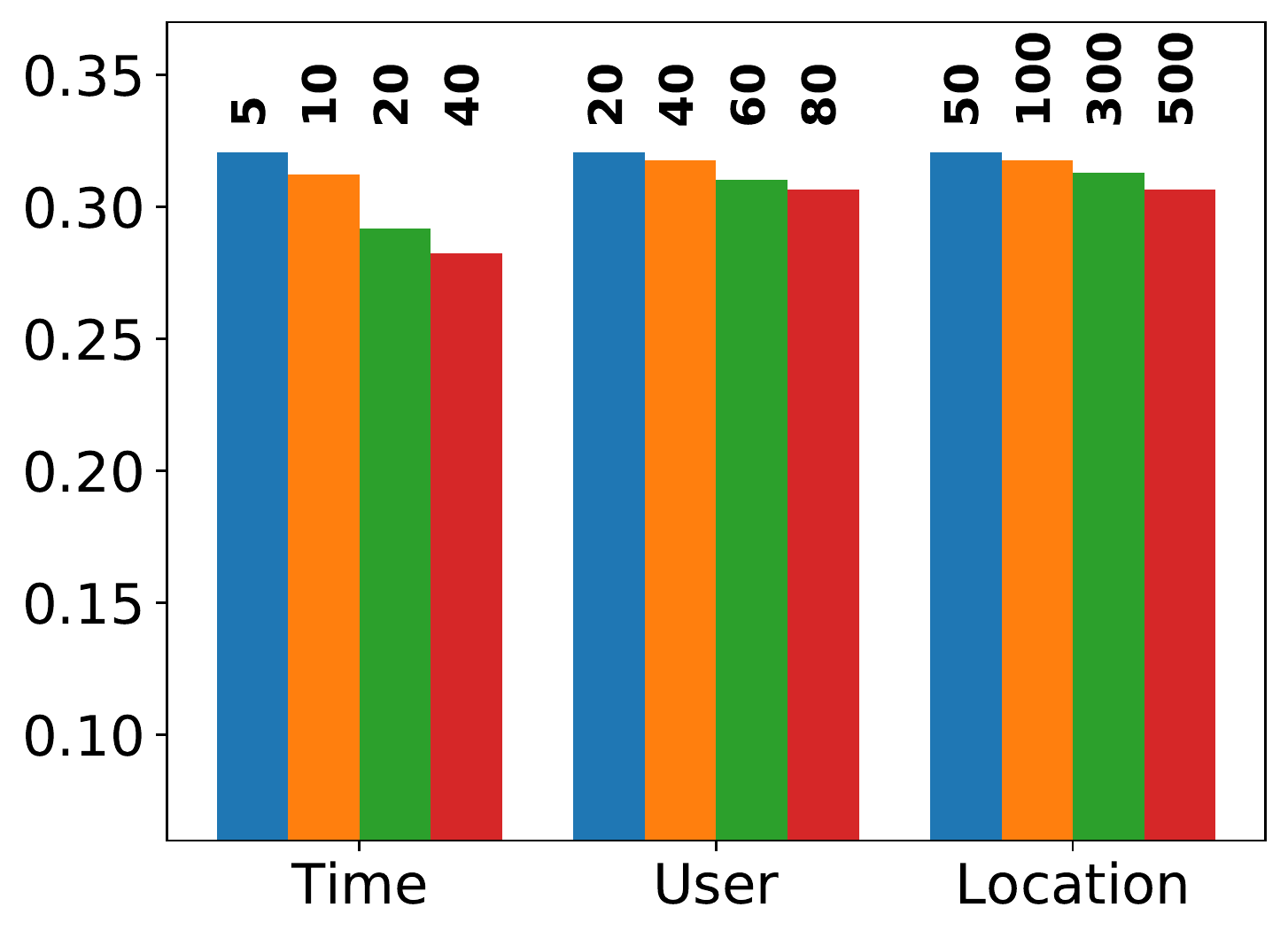}}
\caption{Performance according to embedding size}
\label{fig:ablsize}
\end{figure}

\paragraph{Embedding Size.} We test with various time/user (taxicab)/location embedding sizes in Fig.~\ref{fig:ablsize}. In Taxi, a dimension of 500 for location outperforms others whereas 50 is the best is LA, which is well-aligned with our earlier outcome that LA's user movement patterns are simpler than those of other datasets and simple models work better than sophisticated models with overcapacity for LA.

\subsection{Case Study}

\begin{figure}
\centering
\subfigure[DeepMove]{\includegraphics[width=0.49\columnwidth,trim={0 0 0 0},clip]{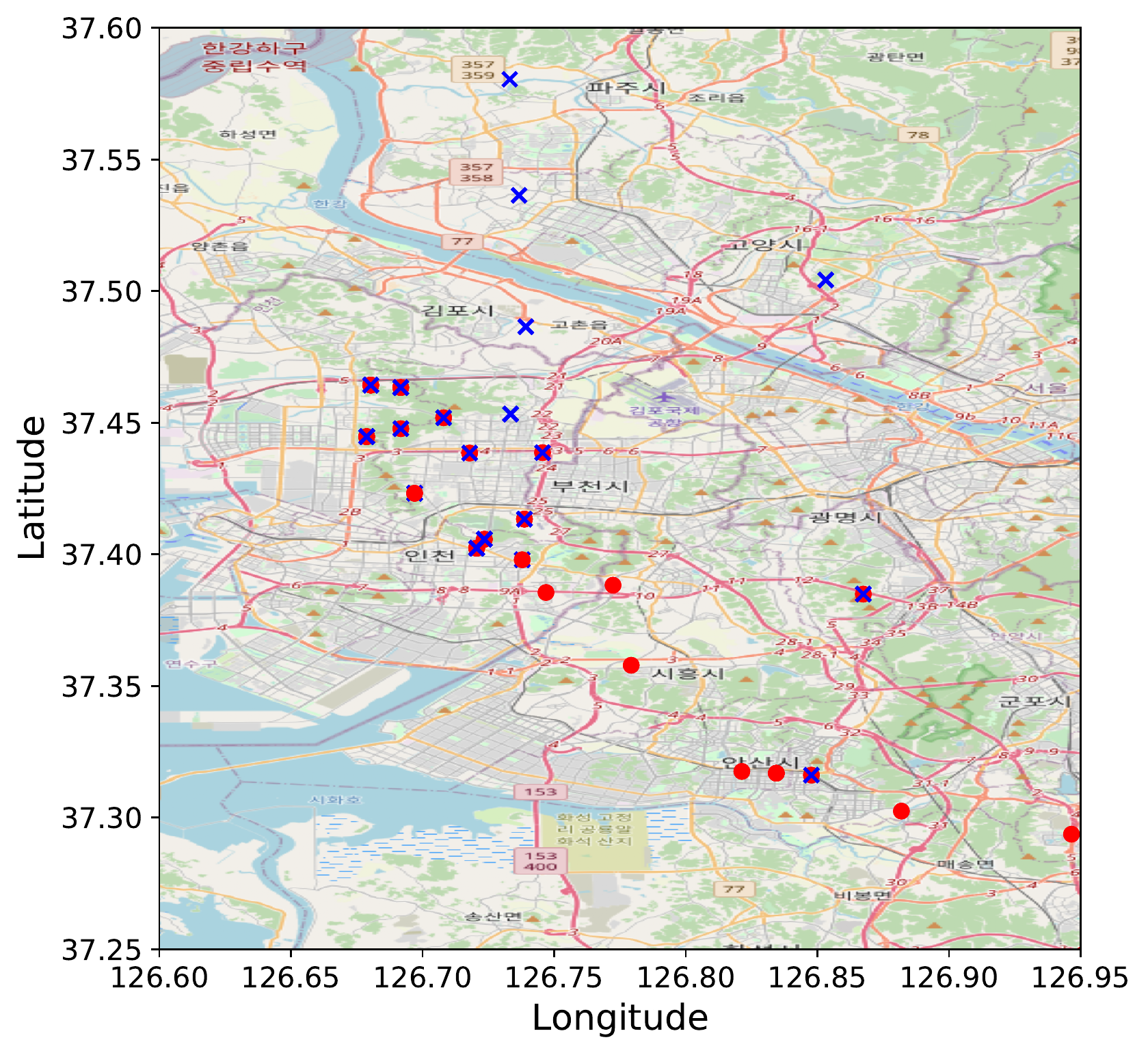}}
\subfigure[LightMove(G2E)]{\includegraphics[width=0.49\columnwidth,trim={0 0 0 0},clip]{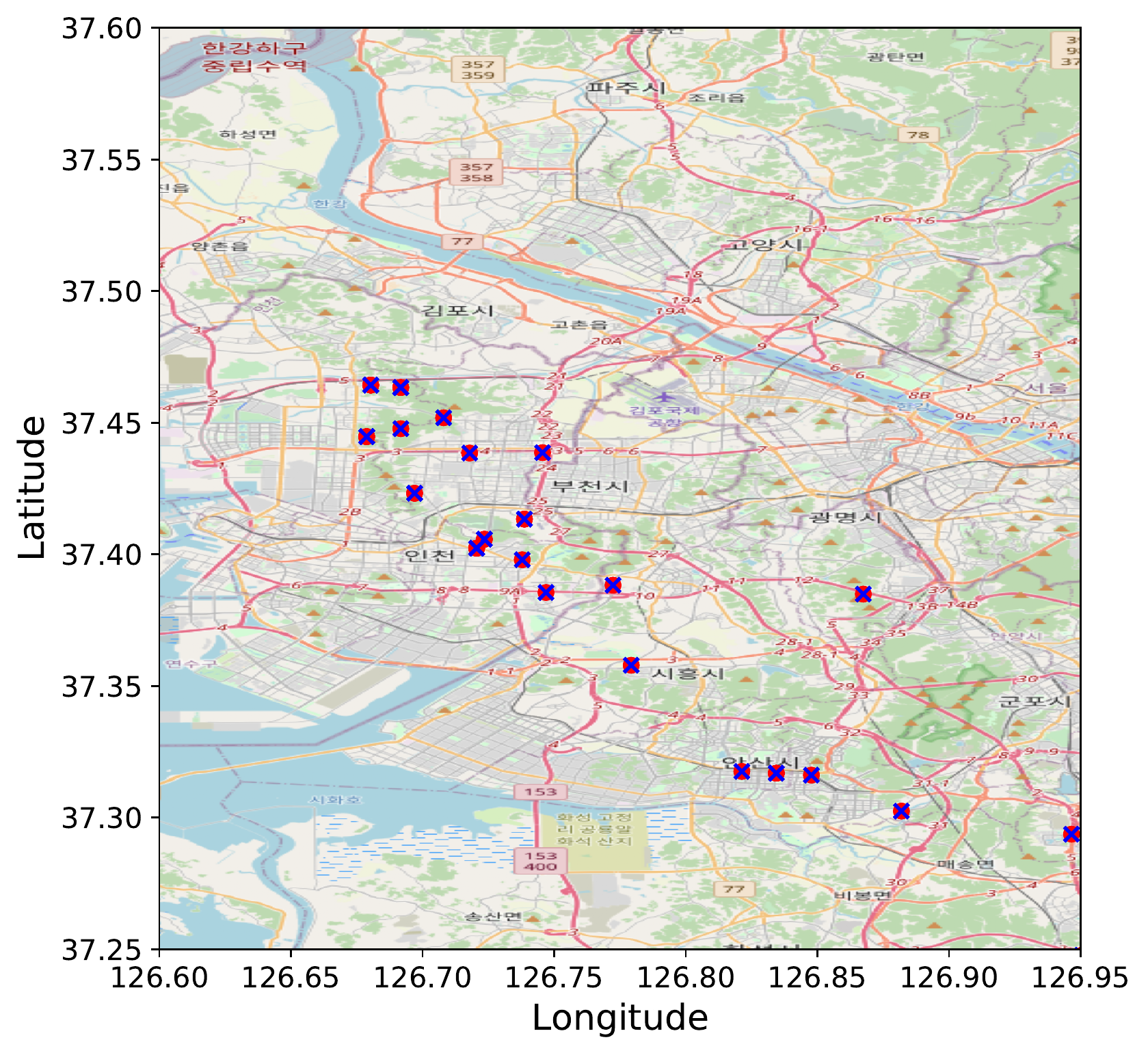}}
\subfigure[DeepMove]{\includegraphics[width=0.49\columnwidth,trim={0 0 0 0},clip]{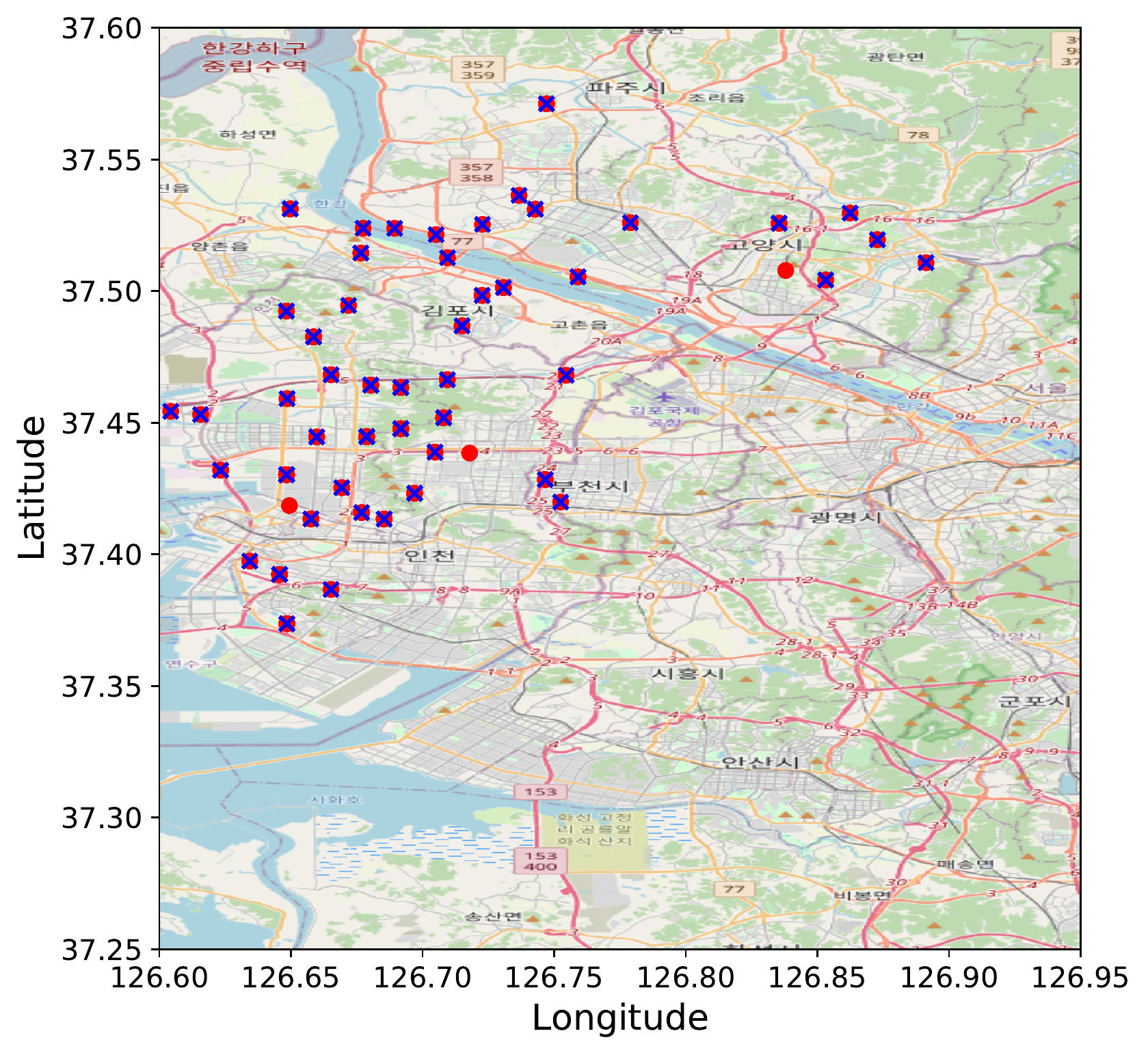}}
\subfigure[LightMove(G2E)]{\includegraphics[width=0.49\columnwidth,trim={0 0 0 0},clip]{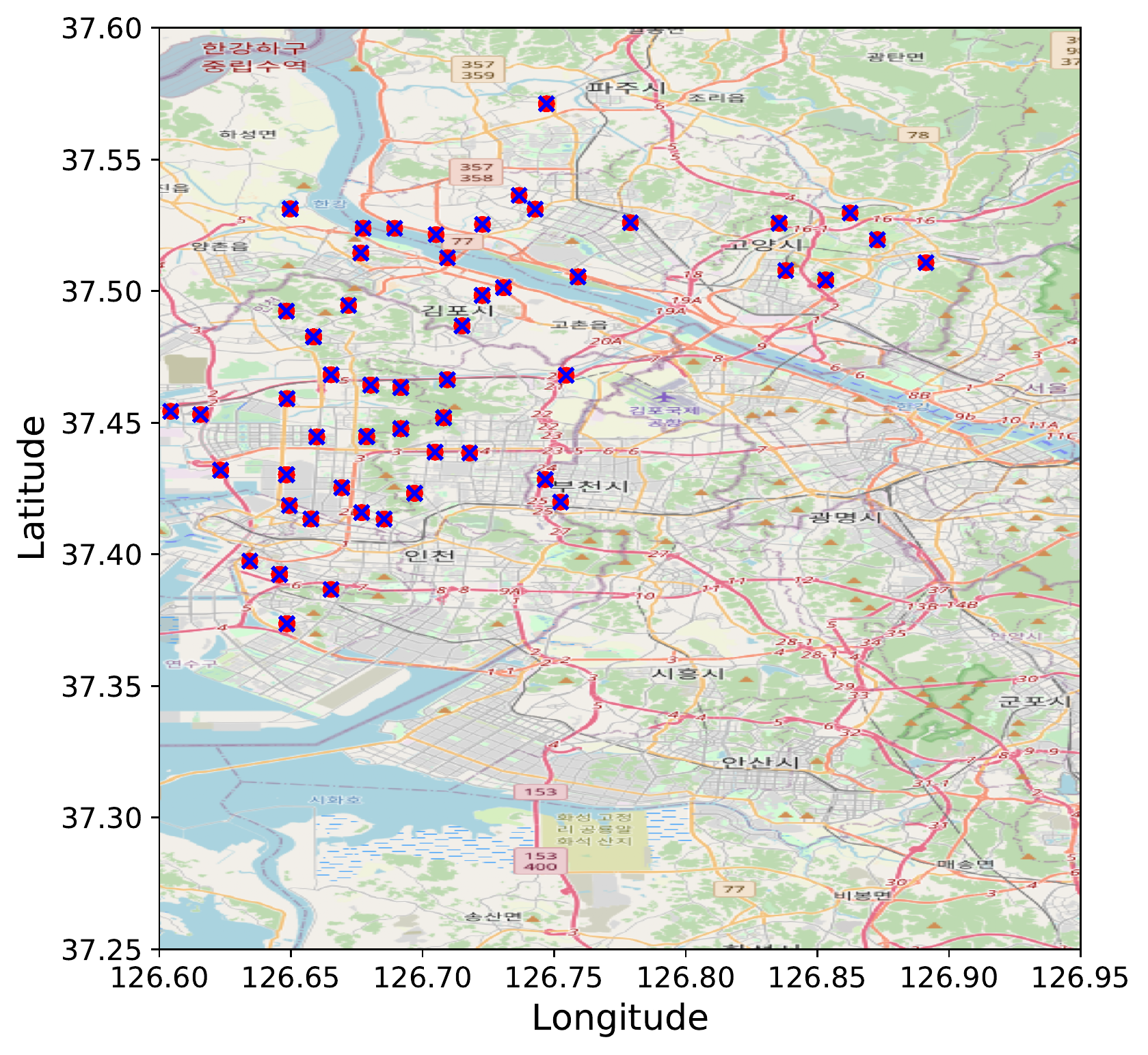}}
\caption{Next location predictions in Taxi by DeepMove and our LightMove. Red means true and blue means predictions.}
\label{fig:case}
\end{figure}
We introduce several visualizations of the prediction examples by our method and DeepMove. In Fig.~\ref{fig:case}, we show the predictions of two test cases. Red circles are true trajectories and blue crosses mean predicted trajectories. In the first test case, a taxicab moves from a corner to another corner. DeepMove does not predict appropriately in such cases. Its predictions say that the taxicab stays around a region whereas our LightMove correctly predicts for most of their locations.

In the last test case, however, both DeepMove and LightMove predict well for most of the locations. In general, DeepMove predicts well when a taxicab stays in a region rather than moving from one corner of the figure to another.

\section{Conclusions}
Taxicab rooftop billboards are effective in increasing brand-awareness according to a recent Nielsen report. For our mobile targeted advertising business, we tackled a practical problem of predicting next-POI locations of taxicabs. We adopt NODEs for designing our lightweight yet accuracy model. Our model shows the best accuracy for our dataset with a relatively faster inference time than state-of-the-art models. Our method also marked the best accuracy in all other standard benchmark datasets. Because we are paid by successful targeted impressions, our task is a key in maintaining our business sustainable.

\begin{acks}
Noseong Park is the corresponding author. This work was supported by the Institute of Information \& Communications Technology Planning \& Evaluation (IITP) grant funded by the Korea government (MSIT) (No. 2020-0-01361, Artificial Intelligence Graduate School Program (Yonsei University)).
\end{acks}

\clearpage
\bibliographystyle{ACM-Reference-Format}
\bibliography{sample-base}

\end{document}